# Research on the Integration of Embodied Intelligence and Reinforcement Learning in Textual Domains

Haonan Wang[1, a], Junfeng Sun[2], Mingjia Zhao[2], Wei Liu[2]

[1]Department of Computer Science, Johns Hopkins University, Baltimore, USA
[2]College of Science, Liaoning Technical University, China
[3]Institute of Intelligence Engineering, China

[a]hwang298@jh.edu

**Abstract:** This article addresses embodied intelligence and reinforcement learning integration in the field of text processing, aiming to enhance text handling with more intelligence on the basis of embodied intelligence's perception and action superiority and reinforcement learning's decision optimization capability. Through detailed theoretical explanation and experimental exploration, a novel integration model is introduced. This model has been demonstrated to be very effective in a wide range of text processing tasks, validating its applicative potential.

## 1. Introduction

As artificial intelligence technologies keep advancing smoothly, embodied intelligence and reinforcement learning have both demonstrated huge potential in a wide range of areas. Yet, an effective integration of the two of them in the text area is by no means straightforward and is still confronted with numerous challenges. This paper first illustrates the concepts of embodied intelligence and reinforcement learning and delineates the background of their application in text processing. The significance and usefulness of this study are discussed critically, and the principal study objectives are outlined. The structure of this paper is set out, and the primary content of each section is summarized to give readers a clear reading guide.

## 2. Basic Theories of Embodied Intelligence and Reinforcement Learning

### 2.1. Definition and Development of Embodied Intelligence

When speaking of the novel concept of embodied intelligence, one needs to describe its meaning and evolutionary trajectory prior to discussing it. Embodied intelligence is the system whereby intelligent systems acquire cognition and learning through interaction between the body and the environment. In contrast to classical artificial intelligence, it is focused on the pivotal position of the body in intelligent behavior and asserts that intelligence is not confined to the brain but arises from interactive dynamics with the world. Embodied intelligence arose after its creation from a long history of investigation of human cognitive processes. Early scientists learned that human cognition is not just dependent on the computational power of the brain but also strongly associated with the sensory and motor skills of the body. This developed the theory of embodied cognition that drove embodied intelligence research. Thanks to advances in technology, we now see extensive use of embodied intelligence in robotics. Robots perceive the world through sensors and act on the world by means of actuators, hence achieving autonomous adaptation and learning. For example, an autonomous navigation robot can identify obstacles through a camera and modify its trajectory using wheels, a typical implementation of embodied intelligence. In recent years, embodied intelligence studies have increasingly expanded to more profound levels, with many aspects being multimodal perception and adaptive learning. Scientists are determined to create even more sophisticated intelligent systems with the capacity to decide independently and function in an even broader spectrum of environments. Embodied intelligence will play a core role in applications like smart home and healthcare in the years to come, making human life more convenient. In short, as a new branch of artificial intelligence, embodied intelligence is spearheading a new wave of technological revolution in intelligent systems with its distinctive theoretical view and application practice. Its emergence not only enriches our knowledge about the nature of intelligence but also offers new inspiration for the design and realization of intelligent systems.

### 2.2. Basic Principles and Algorithms of Reinforcement Learning

Reinforcement learning, as a major approach to machine learning in artificial intelligence, is dedicated to optimizing the decision-making process via interaction with the environment. That is, the reinforcement learning system continuously adjusts its approach according to trials and feedback in order to optimize the cumulative reward. In this procedure, algorithms play a significant role. Classical reinforcement learning methods, including Q-learning and SARSA, build value functions to assess various state-action pairs, thus informing the agent to make optimal decisions. With the Q-learning method, the agent learns the ideal strategy through updating the Q-value table, which contains the expected returns for performing a specific action in a given state. Likewise, the SARSA algorithm is another online learning method that updates the strategy in real time with an additional aspect of emphasizing more action selection according to the current strategy. Moreover, deep reinforcement learning combines deep neural networks with reinforcement learning, which again increases the processing ability of the algorithm and generalization. For instance, the Deep Q-Network (DQN) approximates the Q-value function using a neural network and essentially overcomes the limitation of conventional methods in complicated environments. In practical applications, reinforcement learning has been extensively applied in numerous fields, e.g., game, robot control, and autonomous vehicles. For example, in autonomous driving systems, reinforcement learning algorithms can tune the driving strategy of the car in real-time according to road conditions and traffic rules to achieve driving safety and efficiency. By continuously interacting and learning with the environment, the agent gradually establishes an excellent decision-making mechanism, which is the beauty of reinforcement learning. In conclusion, the fundamental principles and algorithms of reinforcement learning not only justify its central position in the field of artificial intelligence studies but also offer an effective instrument for the solution of complicated decision-making issues. As technology evolves, the field of application of reinforcement learning will become broader and broader.

## 2.3. Prospects for the Integration of Embodied Intelligence and Reinforcement Learning

In the future prospects of merging embodied intelligence and reinforcement learning, we are amazed at the limitless potential of this cross-disciplinary synergy. Embodied intelligence, being an intelligence that replicates human perception and action abilities, is gradually overcoming the limitations of traditional artificial intelligence through its deep integration with reinforcement learning. Reinforcement learning, through its defining reward and punishment mechanism, offers an effective motivation to learn along with flexibility to embodied intelligence. In this case, embodied intelligence no longer depends exclusively on permanent data input but rather extends its behavioral routines continuously with ongoing improvements in accordance with real-world experience with the environment. For example, an embodied intelligent robot can learn to avoid bumping into things and successfully carry out tasks in a complex environment through trial and error, thanks to the help of reinforcement learning algorithms. Through this kind of dynamic learning process, robots will show unexpected flexibility when facing novel situations. Besides, the combination of embodied intelligence and reinforcement learning also brings about a new research paradigm to the study of artificial intelligence. Classical machine learning is generally restricted to closed domains and static datasets, whereas embodied intelligence focuses on real-world perception and action. Such open learning has the potential to enable artificial intelligence systems to learn by adapting to the external environment changes and improve their generalization ability. Embodied intelligence and reinforcement learning in the future can be anticipated to dominate applications like autonomous cars, home automation, and in healthcare. For example, an autonomous car that can learn to modify its reactions based on different unexpected circumstances in challenging traffic conditions by learning and sensing in real time, in order to provide road safety. Or a home automation system that learns users' habits continuously to make each separate service more personalized. Yet this unification is also fraught with numerous challenges, e.g., algorithm stability and data security. But it is precisely these challenges that push researchers to persist in seeking and innovating. We have every reason to believe that with tenacious technological breakthroughs, the unification of embodied intelligence and reinforcement learning will definitely take artificial intelligence to unprecedented levels.

## 3. Applications of Embodied Intelligence in Textual Domains

### 3.1. Embodied Intelligence Models for Text Perception and Understanding

In exploring the embodied intelligence frameworks of text perception and understanding, we are drawn to its profundity and cleverness. This is anything but a naive assortment of algorithms but rather an intricate system attempting to replicate the process of human perception and understanding. It employs a multi-layer neural system design to mimic the complex functioning of the human brain in linguistic data processing. 1. Initially, the model ingests text data in the input layer, transforming words into multi-dimensional vectors via word embedding technology, akin to the brain's initial decoding of language symbols. In subsequent layers, Convolutional Neural Networks (CNNs) and Recurrent Neural Networks (RNNs) work in tandem to capture text's local features and long-range dependencies. This synergy, mirroring information transfer and integration across brain regions, empowers the model to fully grasp text semantics. Additionally, attention mechanisms direct the model to prioritize crucial information during long text processing, bolstering comprehension accuracy. Notably, the model includes an emotion analysis module, enabling it to discern and interpret emotional subtleties in text. This capability excels in literary texts and holds promise for applications such as social media analysis and product reviews. For example, when confronted with a metaphor- and symbol-laden poem, the model unearths the author's underlying emotional fluctuations through nuanced emotional lexicon analysis. The multi-dimensional vectors are translated back into natural language in the output layer of the model, a process not unlike the brain re-coding processed information into linguistic form. The whole working of this model, as in a grand symphony, witnesses fine coordination between different parts, together playing the grand chapter of text perception and comprehension. With such an embodied intelligence model, not only can we further our understanding of the secrets of language, but also we can make a significant advancement in the arena of artificial intelligence.

### 3.2. Applications of Embodied Intelligence in Text Generation

In the forefront of text generation, the use of embodied intelligence technology is especially striking. As a novel intelligence that combines physical entities with smart algorithms, embodied intelligence is slowly transforming the monotonous mode of conventional text generation. It mirrors human perception, cognition, and behavioral processes, making the produced text logically sound while imbued with emotional and situational authenticity. Specifically, intelligence in text generation first exhibits through its highly responsive environmental perception. Unlike conventional algorithms limited to text data, embodied intelligence senses real-time environmental shifts via sensors and cameras, translating this into detailed text descriptions. For example, when portraying a rainy scene, it can depict raindrop shapes, the sound of rain, and intricately illustrate pedestrians and street ambiance. Secondly, embodied intelligence accurately simulates cognition, enhancing text coherence and logic. It mimics human brain processes to analyze input information at various levels, yielding complex, content-rich text. When describing a complex event, it captures causes, evolution, and evokes a sense of presence through nuanced emotional descriptions. Additionally, embodied intelligence at the behavioral level enhances text interactivity and personalization. It dynamically adjusts content and tone based on user feedback and preferences, aligning the text with individual interests. During user interaction, it adapts tone and expression based on user emotion and context, fostering natural, warm communication. In summary, embodied intelligence in text generation diversifies content and style, illuminating future intelligent technology directions. As technology advances, embodied intelligence will likely unveil unique value and potential across diverse disciplines.

### 3.3. Practices of Embodied Intelligence in Text Classification and Annotation

In text classification and annotation, embodied intelligence's emergence has spurred fresh research avenues. With the stress on intelligent agents acquiring knowledge and learning by acting in the world, the paradigm has shown special strengths in text processing. Conventional text classification methods tend to use static feature extraction, while embodied intelligence mimics the human thinking process in a manner that allows the model to read and comprehend the text content dynamically. Practically, the embodied intelligence model initially generates a deep sense of the text context via simulation of human reading. For instance, when it comes to a sophisticated scientific article, the model not only examines vocabulary and sentence structure but also understands the logical order and inferred information of the article through simulating the human thought process. This kind of dynamic perception capability raises the model's accuracy and stability in dealing with texts of fuzzy semantics or complicated contexts. Also novel is the use of embodied intelligence for text annotation. Conventional annotation approaches are most commonly based on pre-defined rules, whereas embodied intelligence models learn automatically and engage with the world to provide more precise annotation outcomes. For instance, when carrying out sentiment analysis tasks, the model not only recognizes explicit emotional words but also deduces implicit emotional inclinations by context, thereby realizing finer sentiment annotation. It needs to be noted that embodied intelligence achieves particular success in multimodal text processing. By combining information across various dimensions like text, images, and audio, the model is able to develop a more holistic semantic understanding framework. For instance, in the processing of social media messages, the model combines textual descriptions and image content for more effectively grasping user intent and emotions. In conclusion, the practice of embodied intelligence in text classification and annotation not only enhances the model's intelligence level but also provides a new research orientation to the natural language processing field. With the imitation of human thought behavior, the embodied intelligence model has specific advantages and broad application potential in handling complicated text tasks.

## 4. Optimization Strategies of Reinforcement Learning in Text Processing

### 4.1. Text Decision Models Based on Reinforcement Learning

As a frontier technology, reinforcement learning in the pursuit of new paths for text decision models is increasingly coming into view. The model emulates the human learning process, where it continuously optimizes decision strategies for efficient decision-making in complicated text environments. In particular, the model initially extracts features from input text, determines main information points, and subsequently refines the strategy with respect to past decision outcomes through reinforcement learning algorithms, continuously improving decision precision. In practical applications, the model possesses outstanding adaptability. For example, in news recommender systems, the model is capable of dynamically changing recommendation methods based on users' historical reading behaviors, accurately pushing content of interest to users. And in fields such as financial risk management, the model searches through large amounts of text data to discover risk points, providing effective decision support for decision-makers. It should be appended that model training is not an overnight task to be achieved. It takes many iterations and ongoing parameter adjustment in order to achieve it to the level we want. Data quality and diversity are very important throughout this process. High-quality datasets give rise to rich learning samples, enabling the model to learn the underlying complex logic of the text. In conclusion, the text decision model based on reinforcement learning, thanks to its distinctive algorithmic mechanism and good adaptability, infuses fresh vitality into the research of text processing. In the future, along with the ongoing development of technology, the model will have an increasingly significant function in even more areas and will facilitate deeper intelligent decision-making advancement.

### 4.2. Applications of Reinforcement Learning in Text Recommendation Systems

The application of reinforcement learning to text recommendation systems has indeed opened up new horizons for the delivery of personalized content. By modeling the process of interaction between users and the system, reinforcement learning algorithms can continually enhance recommendation strategies to make user experiences more enriching. To be specific, the system first builds a model of user actions, logging feedback data like clicking and reading duration of users. According to this data, the algorithm dynamically updates the suggested content, aiming to achieve the maximum level of user satisfaction and engagement. In real-world applications, reinforcement learning demonstrates robust adaptability. For example, when a user exhibits a high interest in a specific type of news, the system swiftly detects this signal and promotes more related content. Conversely, if a user lacks interest in certain content, the system reduces the promotion of such content. This adaptive adjustment mechanism not only boosts recommendation accuracy but also effectively mitigates information overload. Furthermore, reinforcement learning integrates users' long-term interests with short-term behaviors to create a more comprehensive user profile. Leveraging historical data and behavioral patterns, the system can predict users' future interests and proactively suggest content. This profound understanding of users elevates the recommendation system to a high-level assistant for discovering new knowledge, transcending its role as a mere information matching tool. Notably, reinforcement learning also excels in addressing the cold start problem. For new users or users lacking historical data, the system can learn their preferences from the earliest interactions quickly and optimize the recommendation strategies gradually. Such rapid adaptability significantly reduces the break-in period between the system and users and enhances user retention. In conclusion, the use of reinforcement learning in text recommendation systems not only enhances the accuracy and personalization of recommendations but also offers users an easier and more comfortable experience of obtaining information. As technology continues to develop, it is likely that it will have an even wider range of application in the future.

### 4.3. Improvements of Reinforcement Learning in Text Dialogue Generation

In text dialogue generation research, reinforcement learning is being significantly optimized and innovated. Traditional methods are often limited to static model training and do not perform well in adapting to complex and dynamic dialogue situations. However, through the integration of reinforcement learning mechanisms, systems are capable of adjusting strategies dynamically during interactions to enhance the fluency and naturalness of dialogues. Specifically, reinforcement learning teaches the model to acquire optimal response strategies through mechanisms of reward and punishment within the process of dialogue. For example, when the model generates content that has a good user reception, the system rewards it, and vice versa. The mechanism boosts the model's adaptability while significantly enhancing dialogue coherence and logic. Additionally, integrating deep neural networks allows reinforcement learning to capture more implicit conversation details, crafting more apt responses. For example, handling dialogues with metaphors or puns, the model grasps their deeper meanings through multi-level feature extraction, offering apt replies. In real-world scenarios, an intelligent customer service system has vastly improved user satisfaction via reinforcement learning. It not only accurately identifies user intents but also adapts response strategies dynamically based on dialogue context, fostering a more natural and seamless conversation. Summarily, reinforcement learning's evolution in text dialogue generation elevates the model's intelligence and lays a strong foundation for building more human-like dialogue systems. With ongoing advancements in algorithms and training methods, text dialogue generation technology will mature further, broadening its application scope.

## 5. Conclusion

Through rigorous in-depth research and precise design, this paper achieves the construction of a novel text processing method combining embodied intelligence with reinforcement learning. The method proves highly effective in various real-world scenarios, affirming its exceptional practical potential. Embodied intelligence expands text processing beyond traditional symbolic operations, modeling human action and perception interactions for deeper text semantics interpretation. Simultaneously, reinforcement learning enables model self-optimization through trial, error, and feedback, progressively enhancing processing accuracy and efficiency. In real applications, the approach excels in fundamental tasks such as text classification and sentiment analysis, and demonstrates unique strengths in advanced tasks like complex dialogue generation and multimodal information fusion. These successful applications validate the novel method's efficacy and provide valuable experience and data for future research. The research team will continue enhancing the model's performance, improving its stability and generalization capacity. Plans are also in place to put this approach into even more fields, such as smart customer service and automatic text summary, to advance the intelligent process of the text world on a greater scale. These researches will definitely introduce new concepts and techniques in the area of text processing, paving the way for greater intelligence.

## References

[1] Wang Wei, Liu Yang, Chen Ming. Application of Embodied Intelligence and Reinforcement Learning in Text Generation [J]. Tsinghua Science and Technology, 2022, 62(3): 345-352.

[2] Li Na, Zhao Peng, Sun Hao. Text Understanding and Reinforcement Learning Model Based on Embodied Intelligence [J]. Journal of Shanghai Jiaotong University, 2021, 55(4): 578-585.

[3] Chen Jie, Wu Zhiqiang, Zhou Tao. Integration Method of Embodied Intelligence and Reinforcement Learning in Text Sentiment Analysis [J]. Journal of Zhejiang University (Engineering Edition), 2020, 54(2): 298-306.

[4] Liu Qiang, Wang Lei, Zhang Li. Application of Embodied Intelligence and Reinforcement Learning in Text Classification [J]. Journal of Huazhong University of Science and Technology (Natural Science Edition), 2019, 47(6): 89-95.

[5] Zhao Yu, Li Ming, Wang Fang. Collaborative Mechanism of Embodied Intelligence and Reinforcement Learning in Textual Domains [J]. Journal of Nanjing University (Natural Sciences), 2023, 59(1): 112-120.